\documentclass{article}

\usepackage[preprint]{neurips_2022}

\usepackage[utf8]{inputenc} 
\usepackage[T1]{fontenc}    
\usepackage{url}            
\usepackage{booktabs}       
\usepackage{amsfonts}       
\usepackage{nicefrac}      
\usepackage{microtype}      
\usepackage{multirow}
\usepackage{lscape}
\usepackage{graphicx}

\usepackage{algpseudocode}  
\usepackage{amsmath}

\usepackage{amssymb}
\usepackage{amsfonts}
\usepackage{fdsymbol}

\usepackage{gensymb}

\usepackage[usenames]{xcolor}

\usepackage[bookmarksnumbered=true]{hyperref}
\usepackage{url}

\definecolor{cite_color}{HTML}{114083}

\definecolor{link_color}{RGB}{153, 0,0} 
\definecolor{url_color}{RGB}{153, 102,  0}
\definecolor{emp_color}{RGB}{0,0,255}

\hypersetup{
 colorlinks,
 citecolor=cite_color,
 linkcolor=link_color,
 urlcolor=url_color}

\graphicspath{{./figs/}}

\usepackage{framed}
\definecolor{shadecolor}{rgb}{0.94, 0.97, 1.0}

\usepackage{epsfig}
\usepackage{amsmath}
\usepackage{amssymb}
\usepackage{mathrsfs}
\usepackage{graphicx}
\usepackage{subfig}
\usepackage{multirow}
\usepackage{booktabs}

\usepackage{wrapfig}
\usepackage{enumerate}
\usepackage{bm}

\usepackage{tabularx}

\usepackage{natbib}
\bibpunct{(}{)}{;}{a}{,}{,}

\usepackage{mathtools}

\usepackage[vlined, ruled, linesnumbered]{algorithm2e}

\SetCommentSty{mycommfont}
\SetKwInOut{Input}{input}
\SetKwInOut{Output}{output}

 \usepackage{cleveref} 
 \crefname{section}{Section}{Sections}
 \crefname{theorem}{Theorem}{Theorems}
 \crefname{lemma}{Lemma}{Lemmas}
 \crefname{equation}{Equation}{Equations}
 \crefname{proposition}{Proposition}{Propositions}
 \crefname{claim}{Claim}{Claims}
\crefname{appendix}{Appendix}{Appendices}
   \crefname{algorithm}{Algorithm}{Algorithms}
 \crefname{figure}{Figure}{Figures}
 \crefname{table}{Table}{Tables}
 \crefname{remark}{Remark}{Remarks}
 \crefname{definition}{Definition}{Definitions}
 \crefname{equatinon}{Equation}{Equations}
 \crefname{corollary}{Corollary}{Corollaries}

\allowdisplaybreaks

\usepackage{thmtools}
\usepackage{thm-restate}

\let \oldtextcircled \textcircled
\renewcommand{\textcircled}[1]{\oldtextcircled{\footnotesize #1}}

\usepackage{enumitem}
\setlist[itemize]{leftmargin=9mm}

\def \x{\mathbf{x}}
\def \y{\mathbf{y}}

\newtheorem{proposition}{Proposition}

\renewcommand{\mid}{|}

\usepackage{float}

\usepackage{xspace}
\newcommand{\ours}[0]{\texttt{DOMI}\xspace}
\newcommand{\invdann}[0]{\texttt{invDANN}\xspace}

\title{Diversity Boosted Learning for Domain Generalization with Large Number of Domains}

\author{
  Xi Leng\\
  School of Science and Engineering, \\
  The Chinese University of Hong Kong (Shenzhen)\\
  \texttt{221019056@link.cuhk.edu.cn}\\
    \And
  Xiaoying Tang\thanks{Corresponding author}\\
  School of Science and Engineering, \\
  The Chinese University of Hong Kong (Shenzhen)\\
  \texttt{tangxiaoying@cuhk.edu.cn}\\
\And
  Yatao Bian\\
  Tencent AI Lab\\
  \texttt{yataobian@tencent.com}\\
}

\usepackage{etoc}
\etocdepthtag.toc{mtchapter}
\etocsettagdepth{mtchapter}{subsection}
\etocsettagdepth{mtappendix}{none}

\begin{document}

\maketitle

\begin{abstract} 

Machine learning algorithms minimizing the average training loss usually suffer from poor generalization performance due to the greedy exploitation of correlations among the training data, which are not stable under distributional shifts. 
It inspires various works for domain generalization (DG), where a series of methods, such as Causal Matching and FISH, work by pairwise domain operations.
They would need $O(n^2)$ pairwise domain operations with $n$  domains,
where each one is often highly expensive. Moreover, while a common objective in the DG literature is to learn invariant representations against domain-induced spurious correlations, we highlight the importance of mitigating spurious correlations caused by \emph{objects}. Based on the observation that  diversity helps mitigate spurious correlations, we propose a \textbf{D}iversity boosted tw\textbf{O}-level sa\textbf{M}pl\textbf{I}ng framework (\ours) utilizing Determinantal Point Processes (DPPs) to efficiently sample the most informative ones among large number of domains. We show that \ours helps train robust models against spurious correlations from both domain-side and object-side, substantially enhancing the performance of the backbone DG algorithms on rotated MNIST, rotated Fashion MNIST, and iwildcam datasets.

\end{abstract}

\section{Introduction}

Machine learning models are typically trained to minimize the average loss on the training set, such as empirical risk minimization (ERM). The effectiveness of machine learning algorithms with ERM relies on the assumption that the testing and training data are identically drawn from the same distribution, which is known as the IID hypothesis. However, distributional shifts between testing and training data are usually inevitable due to data selection biases or unobserved confounders that widely exist in real-life data. Moreover, the data distribution of the training set is likely to be imbalanced. Certain domains may contain the majority of data samples while other domains are only a small fraction of the training set. Under such circumstances, models trained by minimizing average training loss are prone to sink into spurious correlations and suffer from poor generalization performance. Here spurious correlations refer to misleading heuristics that only work for most training examples but can not generalize to data from other distributions that may appear in the test set. The goal of domain generalization (DG) is to learn a model that can generalize well to unseen data distributions after training on more than one data distributions. Different data distributions are denoted as different domains in DG tasks. For example, an image classifier should be able to discriminate the objects whatever the background of the image is. While lots of methods have been derived to efficiently achieve this goal and show good performances, there are two main drawbacks. 

\textbf{Scalability.} With an unprecedented amount of applicable data nowadays, many datasets contain a tremendous amount of domains, or massive data in each domain, or both. For instance, WILDS \citep{koh2021wilds} is a curated collection of benchmark datasets representing distribution shifts faced in the wild. Among these datasets, some contain thousands of domains and OGB-MolPCBA \citep{hu2020open} contains more than one hundred thousand. Besides WILDS, DrugOOD \citep{ji2022drugood} is an out-of-distribution dataset curator and benchmark for AI-aided drug discovery. Datasets of DrugOOD contain hundreds to tens of thousands of domains. In addition to raw data with abundant domains, domain augmentation, leveraged to improve the robustness of models in DG tasks, can also lead to a significant increase in the number of domains. For example, HRM \citep{liu2021heterogeneous} generates heterogeneous domains to help exclude variant features, favoring invariant learning. Under such circumstances, training on the whole dataset in each epoch is computationally prohibitive, especially for methods training by pairwise operations between domains. For instance, the computational complexity of causal matching in MatchDG \citep{mahajan2021domain} and gradient matching in FISH \citep{shi2021gradient} is $O(n^2)$ with $n$ training  domains. 

\textbf{Objective.} Lots of works in the DG field focus entirely on alleviating or excluding impacts from the domain-side. As mentioned before, a good image classifier should be able to discriminate the objects whatever the background of the image is, and one may naturally aim to exclude the impacts from the background while ignoring that from the objects themselves. A general assumption in the DG field is that data in different domains share some ``stable'' features to form the causal correlations. However, a large branch of studies hold the view that the relationship between these "stable" features and the outputs is \textbf{domain-independent} given certain conditions. While plenty of algorithms have been designed to learn such "stable" features and domain-independent correlations, we show that this objective is insufficient, and a simple counterexample is as follows. We highlight the importance of mitigating spurious  correlations induced from the object-side for training a robust model.  

\begin{figure}[tbp]
\centering
\begin{minipage}{0.49\linewidth}
		\centering
		\includegraphics[width=0.9\linewidth]{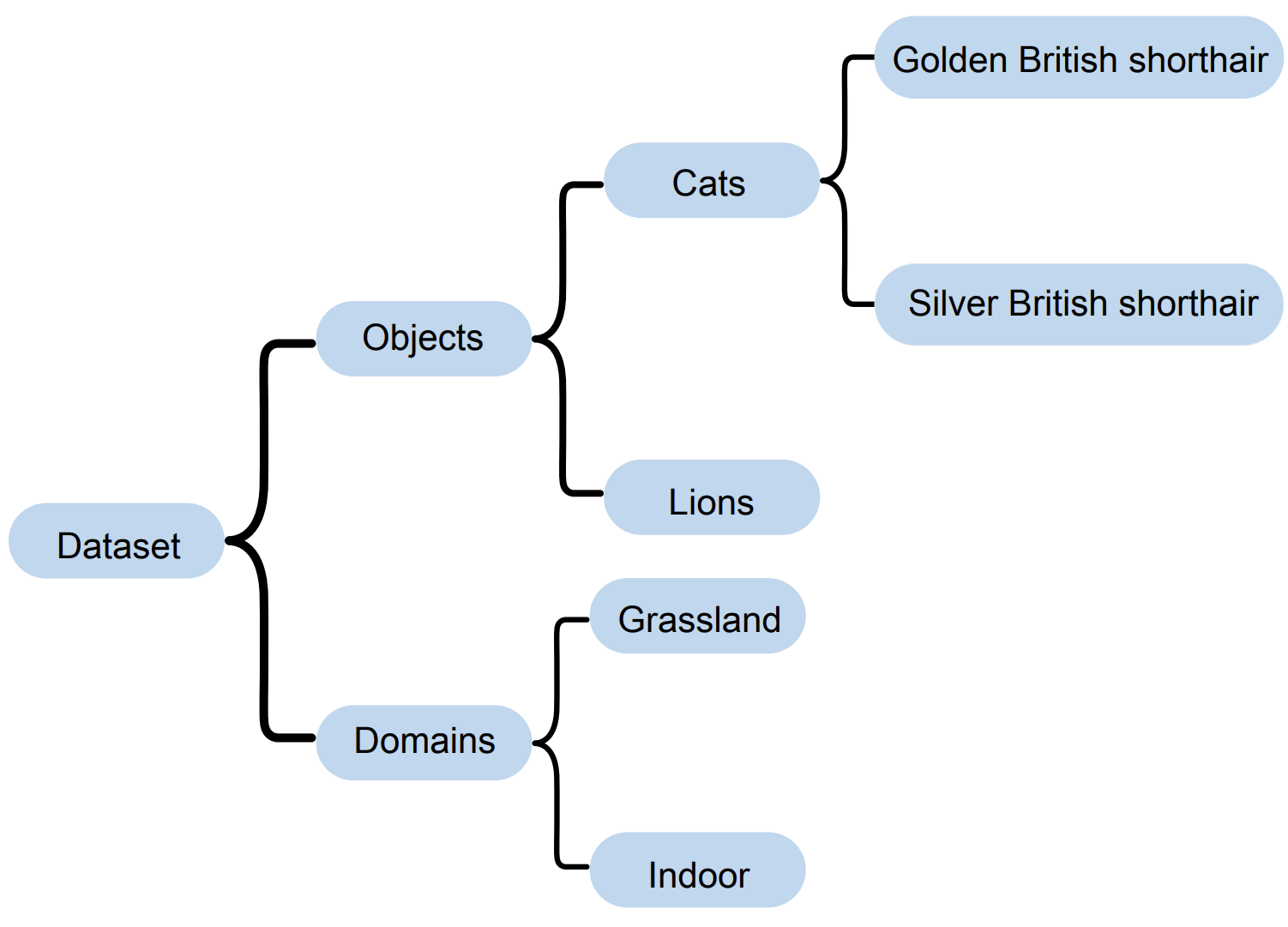}
\end{minipage}
\begin{minipage}{0.49\linewidth}
		\centering
		\includegraphics[width=0.9\linewidth]{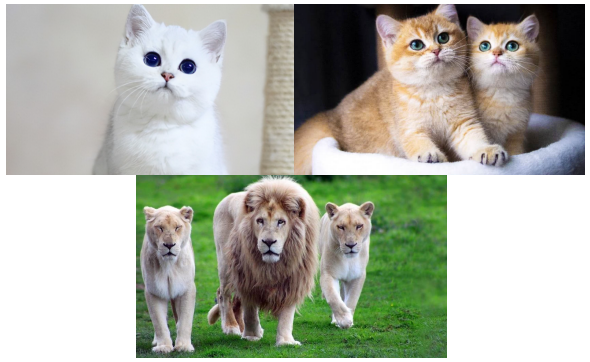}

\hspace{0in}

\end{minipage}
\caption{Dataset of the counterexample. Cats are mainly silver British shorthair (body color of which is silvery white), rarely golden British shorthair (tan), and lions are all tan. As for the background, most of the lions are on the grassland while most of the cats are indoors.}
\label{fig:dataset}
\end{figure}

Suppose our learning task is to train a model to distinguish between cats and lions. The composition of the training dataset is shown in \cref{fig:dataset} and the domain here refers to the background of the figures.

In this example, the correlation between features corresponding to the body color of objects and class labels is undoubtedly independent of domains. Moreover, it helps get high accuracy in the training set by simply taking the tan objects as lions and the white ones as cats. Unfortunately, if this correlation is taken as the causal correlation, the model is prone to poor performance once the distribution of cat breeds shifts in the test set.

To tackle these two issues, a sampling strategy to select the most informative domains or data points for mitigating impacts from both domain-side and object-side to obtain a genuinely robust model is essential. Under the setting of large numbers of domains and domains with massive data points, we propose a diversity boosted two-level sampling framework named \ours. Since we will set forth later that diversity helps mitigate spurious correlations, a sampling scheme to select diverse domains or data points is an essential part of \ours. In this paper, we incorporate Determinantal Point Process (DPP) sampling into \ours as one choice of diversity sampling methods. DPP \citep{kulesza2012determinantal} is a point process that mimics repulsive interactions between samples, and a draw from a DPP yields diversified subsets. Extensive experiments show that \ours helps efficiently alleviate spurious correlations from both domain-side and object-side, substantially enhancing the performance of the backbone DG algorithms on rotated MNIST, rotated Fashion MNIST, and iwildcam.
\paragraph{Summary of contributions.} Our contributions can be summarized  as follows:
\begin{itemize}
\item[] 1. To our best knowledge, this is the first paper to take impacts from the object-side into account for achieving the goal of DG. 
\item[] 2. We propose \ours, a diversity boosted two-level sampling framework to select the most informative domains and data points for mitigating impacts from both domain-side and object-side.
\item[] 3.We show that \ours substantially enhances the test accuracy of the backbone DG algorithms on three benchmarks.
\end{itemize}

\section{Related Work}

\textbf{Domain Generalization.} In DG tasks, the training data is sampled from one or many source domains, while the test data is sampled from the new target domains. The goal of DG is to learn a model that can generalize well to all domains including unseen ones after training on more than one domains \citep{blanchard2011generalizing,wang2022generalizing,zhou2021domain,shen2021towards}. Among recent works on domain generalization, \cite{ben2013robust,duchi2016statistics} utilize distributionally robust optimization (DRO) to minimize the worst-case loss over potential test distributions instead of the average loss of the training data. \cite{sagawa2019distributionally} propose group DRO to train models by minimizing the worst-case loss over groups to avoid learning models relying on spurious correlations and therefore suffering a high loss on some groups of data. \cite{zhai2021doro} further take use of distributional and Outlier Robust Optimization (DORO) to address the problem that DRO is sensitive to outliers and thus suffers from poor performance and severe instability when faced with real, large-scale tasks. On the other hand, \citep{arjovsky2019invariant,javed2020learning,krueger2021out,shi2021invariant,ahuja2020invariant} rather leverage Invariant Risk Minimization (IRM) to learn features inducing invariant optimal predictors over
training environments. However, \cite{rosenfeld2020risks,kamath2021does,ahuja2020empirical} hold the view that works with IRM lack formal guarantees and present analysis to demonstrate IRM fails to generalize well even when faced with some simple data models and fundamentally does not improve over standard ERM. Another branch of studies assume that data from different domains share some “stable” features whose relationships with the outputs are causal correlations and domain-independent given certain conditions \citep{long2015learning,hoffman2018cycada,zhao2018adversarial,zhao2019learning}. Among this branch of work, \cite{li2018domain,ghifary2016scatter,hu2020domain} hold the view that causal correlations
are independent of domain conditioned on class label, and \cite{muandet2013domain} propose DICA to learn representations marginally independent of domain. 

\textbf{MatchDG.} \cite{mahajan2021domain} state that learning representations independent of the domain after conditioning on the class label is insufficient for training a robust model. They propose MatchDG to learn correlations independent of domain conditioned on objects, where objects can be seen as clusters within classes based on similarity. To ensure the learned features are invariant across domains, a term of the distance between each pair of domains is added to the objective to be minimized.

\textbf{FISH.} Different from the two ideas mentioned above, \cite{shi2021gradient} instead propose FISH to achieve the goal of DG. FISH uses an inter-domain gradient matching objective to learn a model with invariant gradient direction in different domains, where the objective augments the ERM loss with an auxiliary term that maximizes the gradient inner product between domains. By minimizing the loss and matching the gradients simultaneously, FISH encourages the optimization paths to be the same for all domains, favoring invariant predictions. To match the gradients while training, FISH incorporates a term of inner product between gradients of each pair of domains into the objective to be maximized.

\textbf{DANN.} \citep{ganin2016domain} incorporates the structure named domain discriminator to implement adversarial training based on the theory that a good classifier for cross-domain shifts should be able to distinguish different classes while cannot learn to identify the domain. \ours takes use of an inverse version of DANN denoted as \invdann to learn domain-side features and help select the most informative domains.

\textbf{DPP.} Determinantal Point Process
(DPP) \citep{kulesza2012determinantal} is a point process that mimics repulsive interactions. Based on a similarity matrix (DPP kernel) of samples to be selected, a draw from a DPP yields diversified subsets. While it shows powerful performance in selecting heterogeneous data, DPP sampling relies on an eigendecomposition of the DPP kernel, whose cubic complexity is a huge impediment. To address this problem, \cite{li2016efficient} suggest to first construct an approximate probability distribution to the true DPP and then efficiently samples from this approximate distribution. As one choice of diversity sampling, DPP sampling is incorporated into \ours to help select the most informative domains and data points, and it can be replaced with other diversity sampling schemes.

Although MatchDG and FISH perform well in domain generalization tasks, the matching procedure between domains means their computational complexity is $O(n^2)$ with $n$ training domains. When $n$ is large, it will inevitably slow down the training process. Therefore, we must select the most informative domains from all the training domains. Inspired by \cite{liu2021heterogeneous} that heterogeneous training domains help to learn invariant features since more variant features can be excluded, we conduct an analysis of diversity and spurious correlations to further state it. To employ DPP sampling to select diverse domains, we build a DPP kernel by measuring the similarity of descriptions of each domain. A description of one domain is derived by a set transformer \citep{lee2019set}. After the featurizer trained by \invdann extracts the features of part or all of data points in a domain, this set of features is transformed to a description.

\section{Diversity Helps Mitigate Spurious Correlations}

Spurious correlations essentially result from the imbalanced data. If a correlation is easy to be found and is held by most of the data, algorithms minimizing the average loss like ERM may simply take this correlation as the causal correlation. Thus when we sample diverse data, we in fact re-balance them and help mitigate spurious correlations. We verify this observation with  a toy example and an experiment.

\subsection{A Toy Example}\label{toy example}

For the task and dataset mentioned above (\cref{fig:dataset}), we further suppose our featurizer extracts 4 features with a binary value as shown in \cref{table:feature}.
\begin{table}[tbp]
\centering
\caption{Details of the features and the label. $X_1$ to $X_3$ correspond to features of the object and $X_4$ corresponds to features of the domain.}
\label{table:feature}
\begin{tabular}{cccccc}
\toprule  
 &  $X_1$ : Mane& $X_2$ : Proportion of face& $X_3$ : Body color & $X_4$ : Background & $y$ \\
\midrule  
0 &  no mane & short face & white & indoors&cat\\
1 &  have a mane & long face & tan & grassland & lion\\

\bottomrule 
\end{tabular}
\end{table}
 
Then $X_1 + X_2 \ge 1 \Rightarrow y=1$ is the causal correlation since the proportion of lions' faces is longer than that of cats, and $X_2$ may be wrongly computed to 0 for male lions because of the existence of mane. $X_3=1 \Rightarrow y=1$ is the Object-spurious-correlation (Abbrev. Osc) and $X_4=1 \Rightarrow y=1$ is the  Domain-spurious-correlation (Abbrev. Dsc). Details of our simulated dataset is shown in \cref{table:feat}.

Suppose we have to get 6 of these 12 data samples for training where 3 of 6 come from cats and another 3 are from lions. There are 4 sampling methods to be picked: random sampling, sampling making the data more diverse on the object features ($X_1$, $X_2$ and $X_3$), sampling making the data more diverse on the domain feature ($X_4$), and sampling making the data more diverse on all 4 features. For convenience, we call these four sampling methods $S_1$ to $S_4$ and use Manhattan Distance on feature tuples to measure the diversity of sampled data. \cref{table:toy} shows the average training accuracy of Osc and Dsc. When the spurious correlations get lower training accuracy, they are more likely to be excluded, favoring exploration of the causal correlations.\par
$S_1$ preserves the imbalance of data. A data point has a larger probability of being sampled into the batch when it appears more often in the dataset. For base-batches sampled by $S_1$, both Osc and Dsc get high accuracy and are thus likely to be wrongly treated as causal correlations.\par
$S_2$ selects data with heterogeneity on object features and data batches sampled by $S_2$ get lower accuracy for Osc than base-batches, which means $S_2$ reduces the probability of taking Osc as causal correlation. However, data batches sampled by $S_2$ get almost the same result for Dsc.\par
$S_3$ selects data with heterogeneity on domain-feature $X_4$. For these batches of data, Dsc gets lower accuracy than base-batches and is less likely to be taken as causal correlation while Osc has a similar result.\par
$S_4$ selects data with heterogeneity on all 4 features. Compared to base-batches, the data batches got by $S_4$ have lower accuracy on both spurious correlations.

\begin{table}[tbp]
\centering
\caption{We use each sampling method to select 30 batches of data and compute average accuracy on two kinds of spurious correlations, which are more likely to be excluded when they get lower accuracy.}
\label{table:toy}
\begin{tabular}{cccc}
\toprule  
Sampling Method&  Accuracy of Osc& Accuracy of Dsc \\
\midrule  
$S_1$&  0.86& 0.68\\
$S_2$&  0.77& 0.66\\
$S_3$&  0.85& 0.50\\
$S_4$& 0.78& 0.49\\
\bottomrule 
\end{tabular}
\end{table}

\subsection{An Empirical Study on Diversity and Spurious Correlations}\label{exp}

Suppose the idea holds, i.e., training on diverse data about certain features can help exclude spurious correlations caused by these features. And now we have two settings for the experiment. One is training models on randomly sampled domain lists for each round and getting their test accuracy. For the other one, with a randomly sampled domain list as the initial domain list, every next domain list is selected by DPP based on the current model. Then the accuracy of the later setting should show a more oscillatory trend.

Why? Suppose we have a model with good generalization ability now, that means the model extracts the "stable" features and knows little about spurious correlations. If we use DPP to sample domains based on this featurizer, we just get diverse data about the "stable" features while they are still imbalanced as to spurious features. Trained on these data, the next model is likely to be affected by spurious correlations and show poor generalization ability. Similarly, suppose we have a model with poor generalization ability. In this case, the model actually learns spurious correlations and data attained by DPP using this featurizer is diverse as to spurious features. Trained on these data, the next model is less likely to be affected by spurious correlations and prone to good generalization ability.

We run experiments under the two settings on the Rotated Fashion MNIST dataset. The rotated degree of the training dataset is from 15 to 75 while that of the test set is 0 and 90. The rotated degree is taken to be domain labels, i.e., the training set gets 61 domains. We denote the former setting, i.e., randomly sampling domains in each round as baseline. About DPP-line, after a featurizer processing all the data of a single domain, we average all outputs and take it as the description of this domain. Then we take use of descriptions of all training domains and cosine similarity measurement to get a similarity matrix. Finally DPP selects domains based on this similarity matrix. We observe the test accuracy of 20 rounds in each setting and compute the variance every time. After repeating 10 times, the average variance in the baseline is 31.3, much lower than the other setting, 82.2, which is in line with our expectation.

\section{An Empirical Study on The Insufficient Objective}

Still on the Rotated Fashion MNIST dataset, in this experiment we train two models with all else being equal on two different domain lists containing five domains each.

How do we obtain these two domain lists? First, we randomly sample a domain list, and use DANN and \invdann(details in \cref{invdann}) to respectively train a featurizer on this domain list for the same epochs. Now we get two featurizers. To differentiate them, we call one object-featurizer and the other domain-featurizer since the former learns representations about object and the latter about domain. Then we use DPP to sample domains based on each featurizer.

Now we get these two domain lists, on which we train models and observe their sensitivity to domain shifts and test accuracy. As for sensitivity, we get descriptions of all 61 domains and compute  similarities between each pair of them just like in DPP procedure. Then we simply compute the sum of these similarities. A smaller summation means more dissimilar among domains and more sensitive to domain shifts, thus depending more on Dsc. The result shows that while the model trained on the domain list obtained by domain-featurizer gets a larger sum (1382.0 compared to 1201.8), it still gets lower test accuracy. This means although training on domains derived by \invdann helps mitigate Dsc, there exists another spurious correlation, i.e., Osc, affecting the generalization ability of models. One possible explanation for the result is: While the two featurizers are trained for the same epochs, the task of differentiating domains (rotation) is more straightforward than identifying images. Thus the object-featurizer is likely to learn both Dsc and Osc because of insufficient learning epochs while the domain-featurizer learns well and only extracts domain features. As is shown in \cref{exp}, data attained by DPP based on object-featurizer is diverse for two kinds of spurious features, training on which helps alleviate two kinds of spurious correlations, and thus gets the model with better test accuracy.

\section{Formulation of Two Main Observations}

\begin{proposition} \label{lemma1}
Diverse domains of data help exclude spurious correlation.
\end{proposition}

Consider a dataset $D = \{D^d\}$ which is a mixture of data $D^d =\{(x_i^d, y_i^d)\}_{i=1}^{n_d}$ where $d$ is one domain of the ground set $D$, $ \x_i^d$ and $\y_i^d$ are the $i_{th}$ data and label from domain $d$ respectively, and $n_d$ is the number of samples in $d$. Suppose we now have dataset $D_k$ consisting of $k$ domains. On $D_k$, the distribution of data is $P^{k}$(X,Y). A "good" set denoted by $C_k$ is a set containing "good" correlations that get high accuracy on $D_k$. The set of causal correlations is $C$. $C \subseteq C_k$ since causal correlations can definitely get good performance but "good" correlations for the k domains may not be held in other domains, i.e., spurious correlations. Our goal is to exclude as many spurious correlations as possible. \\
Given another domain $d_{k+1}$ to form dataset $D_{k+1}$ together with the former k domains. The corresponding data distribution and the "good" set are $P^{k+1}$(X,Y) and $C_{k+1}$, respectively. If $P^{k+1}$(X,Y) is close to $P^{k}$(X,Y), then most of the correlations in $C_k$ will still be "good" for $D_{k+1}$ and thus preserved in $C_{k+1}$. Nevertheless, if $d_{k+1}$ is a heterogeneous domain that can significantly change the distribution of data, then the "good" set after being constrained would be obviously smaller than the original one, i.e., $\mid C_{k+1} \mid < < \mid C_{k} \mid$, showing that diverse domains help exclude spurious correlations and training on which helps obtain robust models.

\begin{proposition} \label{lemma2}
Excluding domain-spurious-correlations is insufficient for learning a robust model.
\end{proposition}

\begin{figure}[tbp]
\centering
\includegraphics[width=6cm]{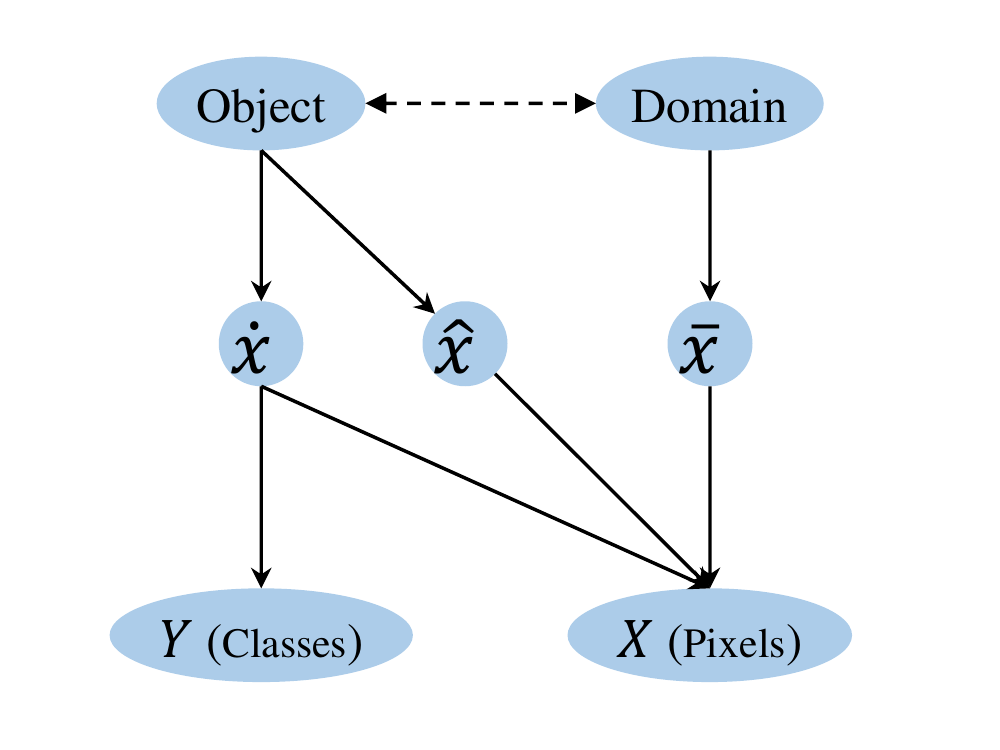}
\caption{The Structural Causal Model for the data-generating process with a node $\hat{\x}$ leading to object-induced spurious correlations.}
\label{fig:structure}
\end{figure}

\cite{mahajan2021domain} have proposed that correlations independent of domain conditional on class $(\Phi(x) \Vbar {D} \mid Y)$ are not necessarily causal correlations if $P(\dot{x}\mid Y)$ changes across domains. Here $\Phi(x)$ is a featurizer to extract features and $\dot{x}$ represents the causal features. We now further propose that the condition is still not sufficient even if $\dot{x}$ is consistent across domains. Since a correlation incorporating features entirely from the objects can also be a spurious correlation. \cref{fig:structure} shows a  structural causal model (SCM) that describes the data-generating process for the domain generalization task.

In this figure, data is divided into two parts: domain-side and object-side. $\overline{x}$ of domain-side is the reason for Dsc. For object-side, feature is further divided into $\dot{x}$ and $\widehat{x}$ where $\widehat{x}$ is the reason for Osc, just like the body color of objects in \cref{toy example}. The three parts together make up the observed data. Thus even if we exclude all the domain-induced spurious correlations, i.e., entirely remove the effect from $\overline{x}$, we may still obtain object-induced spurious correlations resulting from $\widehat{x}$.

\section{Methods}
To sample heterogeneous domains, a powerful method is DPP sampling, a point process which mimics repulsive interactions between samples. Based on the similarity matrix between the data points, a draw from a DPP yields diversified subsets. Using DPP, we propose a diversity boosted two-level sampling framework named \ours to tackle the issue of scalability and help train a robust model by excluding spurious correlations from both the domain-side and object-side.

\subsection{\invdann}\label{invdann}
We utilize \invdann to learn domain representations. In level-one-sampling of \ours, \invdann trains a featurizer to extract domain features and thus help select diverse domains.

Domain-Adversarial Neural Networks (DANN) proposed by \citep{ganin2016domain} is composed by Featurizer, Classifier and Discriminator. Featurizer extracts features of data samples, Classifier learns to classify class labels of data and Discriminator learns to discriminate domains. Since DANN aims to obtain a model can not differentiate domains to ensure Featurizer captures domain-independent features, Discriminator is connected to the Featurizer via a gradient reversal layer that multiplies the gradient by a certain negative constant during backpropagation. Gradient reversal ensures that the feature distributions over the two domains are made similar, thus resulting in domain-independent features.

Using the architecture of DANN, we let Classifier learn to classify domain labels of data while Discriminator learns to discriminate class labels. As an inverse version of DANN, \invdann aims to train a model which can classify domains while can not distinguish class labels. Thus we can get Featurizer extracting only domain-side features.

\subsection{\ours}
\begin{figure}[tbp]
\centering
\includegraphics[width=12cm]{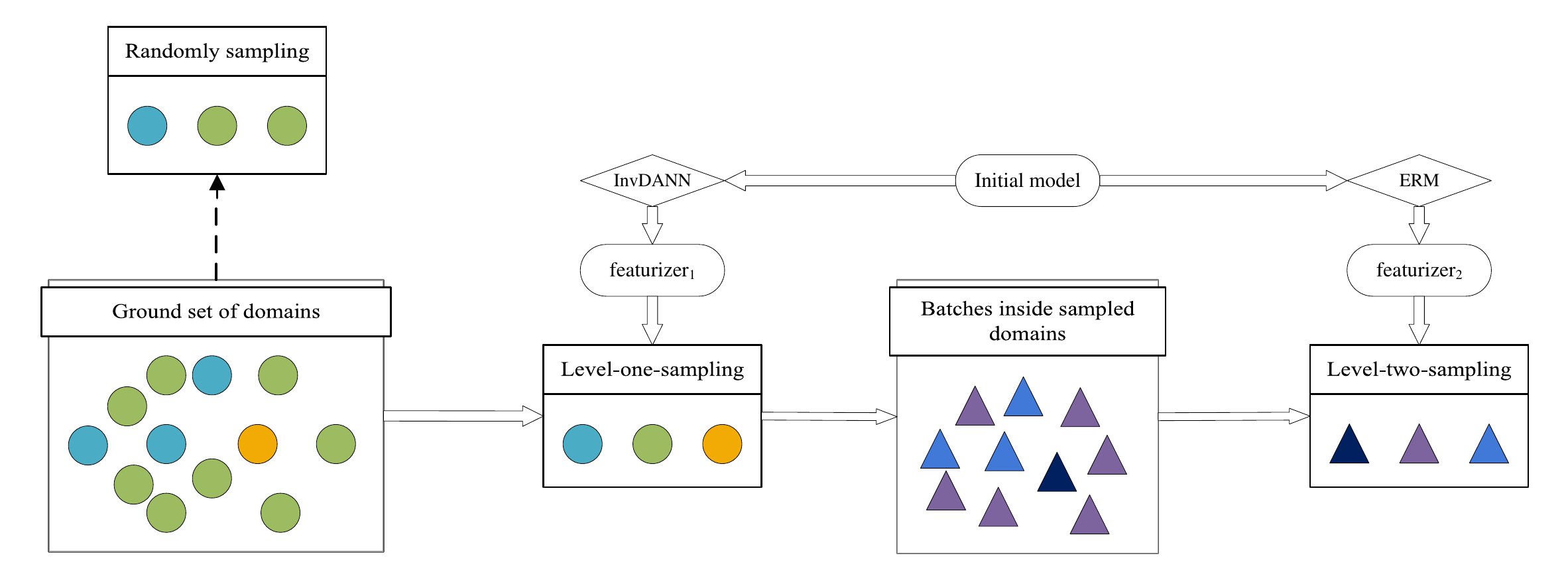}
\caption{Illustration of the sampling framework of \ours. The solid arrow indicates the actual sampling flow, while the dotted arrow only indicates the difference between randomly sampling and \ours.}

\label{fig:domi}
\end{figure}

\cref{fig:domi} shows the sampling procedure of \ours. In level-one-sampling of \ours, we first use \invdann to train a featurizer extracting features of domains rather than objects on a subset of domains and data. Since the featurizer aims at domain-side features, DPP based on $L_d$ can select domains to help exclude domain-induced spurious correlations according to \cref{lemma1}. Then, in level-two-sampling, since we do not have available labels just like domain labels in level one, it is infeasible to utilize \invdann again to train a featurizer. So we instead use ERM to train a featurizer. As we mentioned before, ERM is prone to taking shortcuts and learning spurious correlations. Moreover, since domains attained by level-one-sampling contain diverse data on the domain-side, ERM can avert learning Dscs. Combining these two, ERM in level two can train a featurizer extracting features of Osc. Thus batches selected by DPP based on $L_b$ can help exclude object-induced spurious correlations. Finally, we get a subset of the dataset to tackle the issue of scalability under the setting of tremendous domains and training on which help obtain robust models against impacts from both Osc and Dsc.

\begin{algorithm}[tbp]
  \caption{Sampling Procedure of \ours}
  \label{alg1}
  \KwIn{Dataset $\{(x_i^d, y_i^d)\}_{i=1}^{n_d}$ from $\mid \textbf{D} \mid$ domains}
  \KwOut{Heterogeneous sub-dataset for training}
  \textbf{Level-one-sampling}\\
  Train $featurizer_1$ using \invdann on  $\{(x_i^d, y_i^d)\}_{i=1}^{n < n_d}$ from domain set $D$, $\mid D \mid < \mid \textbf{D} \mid$ \;
  \For{d in $D$}
  {
    Extract features of all data points in $d$ denoted as $feat_d$ by $featurizer_1$ \;
	Taking average of $feat_d$ as the description of $d$\;
  }
  Computing similarity matrix $(L_d)$ of descriptions of all domains\;
  Obtain set of diverse domains $(\Omega)$ by DPP sampling based on $L_d$\;
  \textbf{Level-two-sampling}\\
  Train $featurizer_2$ using \textbf{ERM} on dataset $\{(x_i^d, y_i^d)\}_{i=1}^{n_d}$ from $\Omega$\;
  \For {$b$ in all batches of dataset}
  {
    Extract features of all data points in $b$ denoted as $feat_b$ by $featurizer_2$\;
    Taking average of $feat_b$ as the description of $b$\;
  }
  Computing similarity matrix $(L_b)$ of descriptions of all batches\;
  Obtain diverse batches by DPP sampling based on $L_b$\;
\end{algorithm}

\section{Experiments}

We have investigated the performance of \ours with three backbone DG algorithms on two simulated benchmarks (Rotated MNIST, Rotated Fashion MNIST) and iwildcam, which show that \ours can help substantially get higher test accuracy. The settings and results are shown as follows.

\subsection{Configurations}

\paragraph{Datasets} To satisfy the setting of large number of domains, we extend the original simulated benchmarks on MNIST and Fashion MNIST by \cite{piratla2020efficient} from rotating images 15\degree $ $ through 75\degree $ $ in intervals of 15\degree $ $ to intervals of 1\degree $ $ in the training set, i.e., 61 domains in total. And we get test accuracy on the test set which rotates images either 0\degree $ $ or 90\degree. Moreover, while the original datasets rotate the same images for different degrees, we extend them by rotating different images for each domain to fit the real cases in DG tasks.
WILDS \citep{koh2021wilds} is a curated collection of benchmark datasets representing distribution shifts faced in the wild. As one dataset in WILDS, iwildcam contains photos of wild animals and 324 different camera traps are taken as domains.

\paragraph{Backbones} We take MatchDG \citep{mahajan2021domain}, FISH \citep{shi2021gradient} and DANN \citep{ganin2016domain} as backbone algorithms. The former two algorithms train models by pairwise domain operations, both causal matching of MatchDG and gradient matching of FISH lead to $O(n^2)$ computational complexity with $n$ domains. Since they are prohibitive to be scaled to DG tasks with a large number of domains, it's essential to sample part of the most informative domains. And we further incorporate DANN as one of the backbone algorithms in that \ours can not only efficiently select domains by its first level but can help deal with circumstances where each domain contains massive data by the second level.

\paragraph{Baselines} For each one of the backbone algorithms, we set the baseline as training on domains selected by randomly sampling scheme, compared to level-one-sampling of \ours and complete \ours. We sample 5 domains for training on Rotated MNIST and Fashion MNIST and 10 domains on iwildcam as in the original experiment of \cite{shi2021gradient}. We keep other factors such as model architecture and learning rates the same for different sampling schemes. On iwildcam, we access 30 domains and all data points in each domain for the training of level-one-sampling in \invdann. On the other two datasets, we utilize 40 domains and 750 of 2000 data points in each domain. The number of batches ($\delta$) selected in level-two-sampling is a hyperparameter. A smaller $\delta$ help efficiently mitigate strong Osc and speed up training, but when Osc is weak, a small $\delta$ leads to a waste of training data. When $\delta$ equals the number of entire batches, \ours reduces to \ours with only level one sampling. In the experiment we set $\delta$ as 115 of 157 for Rotated Fashion MNIST and 135 of 157 for Rotated MNIST. On iwildcam, since the number of batches vary among domains, we set $\delta$ as 80\% of total batches in the selected domains. 

\paragraph{Model selection}\label{model-selection}
During training, we use a validation set to measure the model's performance. The test accuracy of the model is updated after an epoch if it shows better validating performance. That is, we save the model showing the highest validation accuracy after the training procedure, obtain its test accuracy and report results. For iwildcam, we use the standard validation set in WILDS. For Rotated MNIST and Fashion MNIST, data from only source domains (rotation degree is from 15 \degree $ $ to 75 \degree $ $) are used to form the validation set since using data from target
domains (rotation degree is 0 \degree $ $ and 90 \degree $ $) for validation goes against the motivation of generalization to unseen domains.

\subsection{MatchDG}\label{matchdg}

MatchDG is a two-phase method, and in our experiment we set 30 epochs of training for phase 1 and 25 epochs for phase 2. We repeat the experiment of MatchDG 20 times with random seeds, and \cref{table:matchdg} shows the average test accuracy of three sampling schemes on two datasets. 

While $level_1$ gets higher accuracy on Rotated MNIST and $level_2$ shows better performance on Fashion MNIST, they all outperform $level_0$, i.e., randomly sampling. 

\begin{table}[tbp]
\centering
\caption{Average test accuracy of MatchDG. Here $level_0$ denotes randomly sampling, $level_1$ denotes level-one-sampling of \ours and $level_2$ is the complete version of \ours .}
\label{table:matchdg}
\begin{tabular}{cccc}
\toprule 
& $level_0$& $level_1$& $level_2$ \\
\midrule  
Rotated MNIST& 82.0& \textbf{84.3}& 83.9\\
Fashion MNIST& 38.6&39.5 & \textbf{39.8}\\

\bottomrule 
\end{tabular}
\end{table}

\begin{figure}[tbp]

\begin{minipage}[b]{.56\linewidth}
    \centering
    \subfloat[][Fashion MNIST]{\label{md_fm}\includegraphics[width=7.5cm]{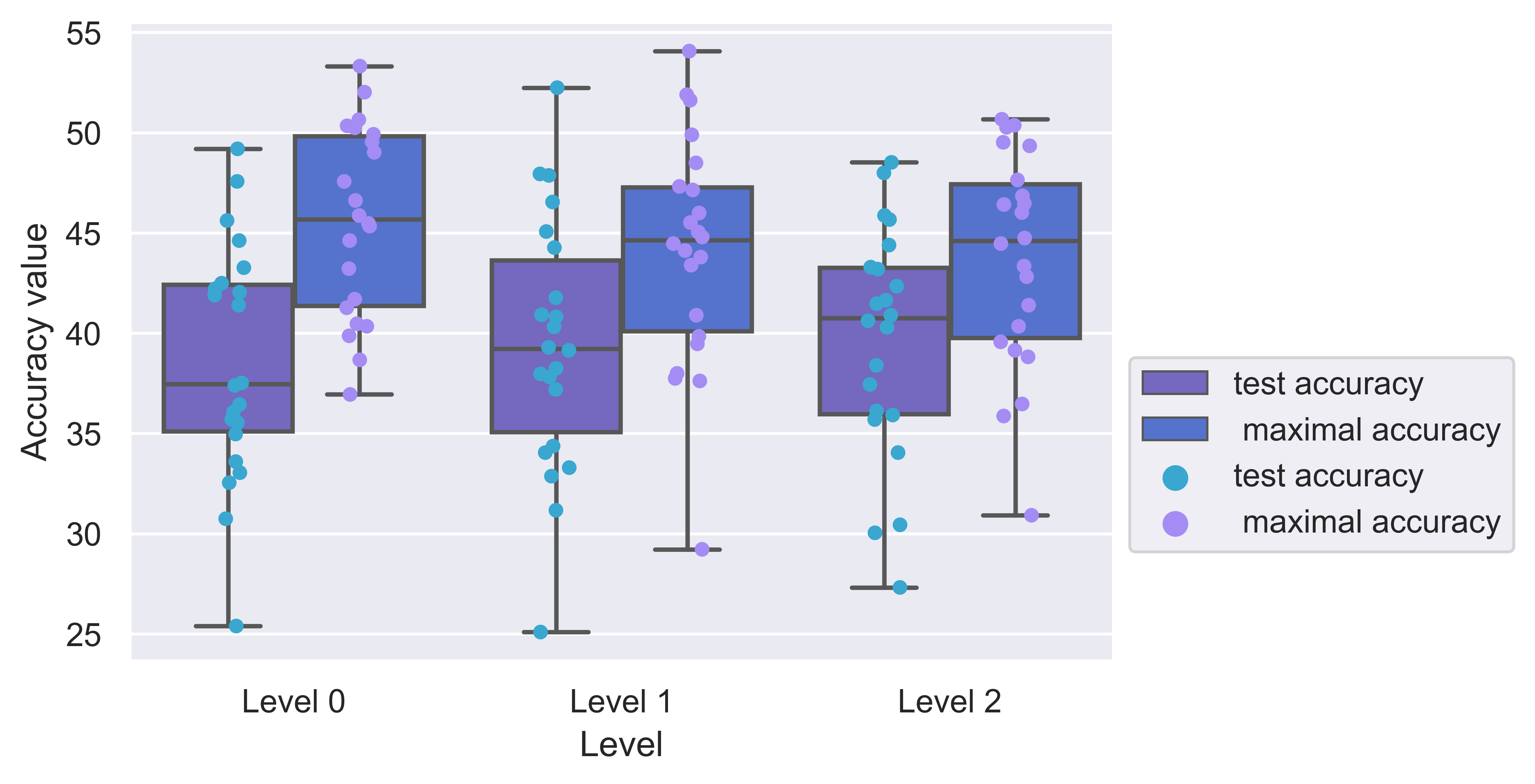}}

\end{minipage} 
\begin{minipage}[b]{.4\linewidth}
    \centering
    \subfloat[][Rotated MNIST]{\label{md_rm}\includegraphics[width=5.5cm]{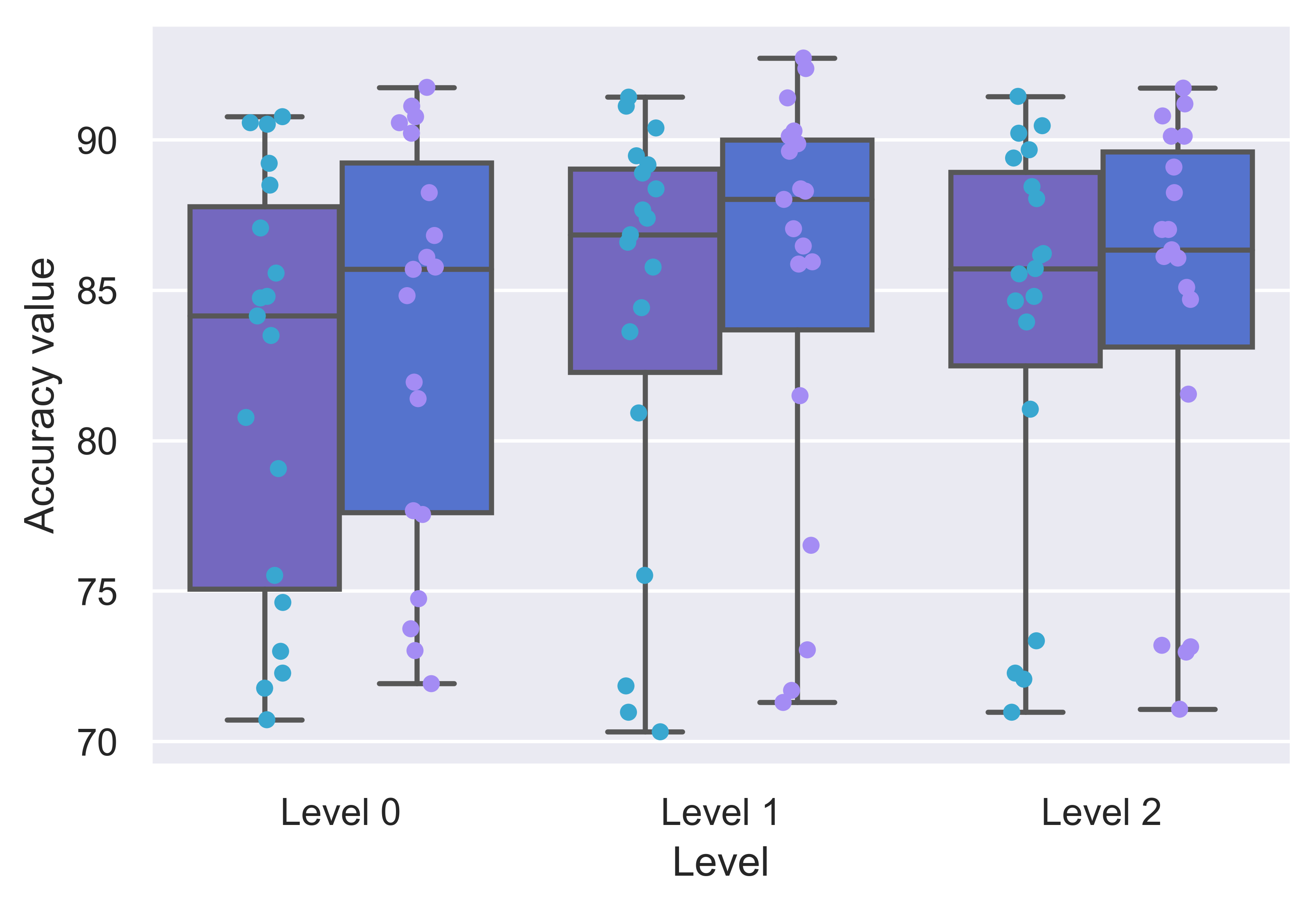}}
\end{minipage}
\caption{Boxplot of test accuracy and maximal accuracy among about 20 times of repeated experiments leveraging different sampling levels on Fashion MNIST and Rotated MNIST. Here each tiny circle represents one experiment, of which the vertical location corresponds to the accuracy value. The horizontal line inside each box indicates the mean value.}
\label{fig:md}
\end{figure}

Moreover, during training we observed that the test accuracy first rises to the peak and then begins to decline along with the increase of validation accuracy. This reduction indicates the model overfits to spurious correlations. Thus we further record the peek value of the test accuracy in each experiment with random seeds and denote it as maximal accuracy. After repeating the experiment about 20 times (19 times on Rotated MNIST and 22 times on Fashion MNIST), the distribution of the test accuracy and maximal accuracy under different sampling schemes is shown in \cref{fig:md}. While the test accuracy of $level_0$ scatters, that of $level_1$ and $level_2$ centers. Moreover, the gap between test accuracy and maximal accuracy of $level_1$ is smaller than that of $level_0$ and $level_2$ further shrinks the gap. \cref{table:matchdg} and \cref{fig:md} indicate that \ours helps train a robust model with good performance.

\subsection{DANN}
\begin{figure}[tbp]

\begin{minipage}[b]{.48\linewidth}
    \centering
    \subfloat[][Fashion MNIST]{\label{dann_fm}\includegraphics[width=6cm]{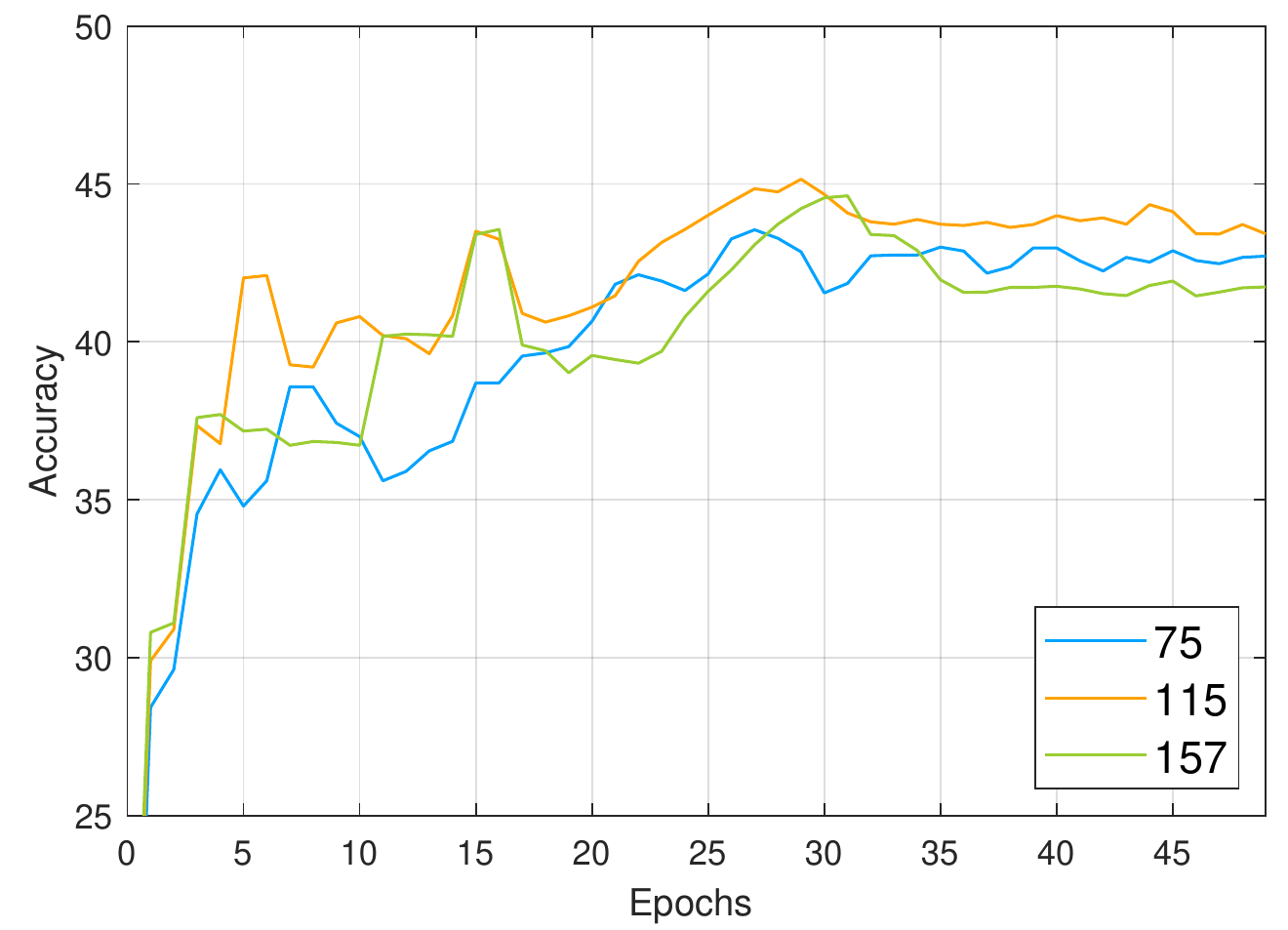}}

\end{minipage} 
\begin{minipage}[b]{.48\linewidth}
    \centering
    \subfloat[][Rotated MNIST]{\label{dann_rm}\includegraphics[width=6cm]{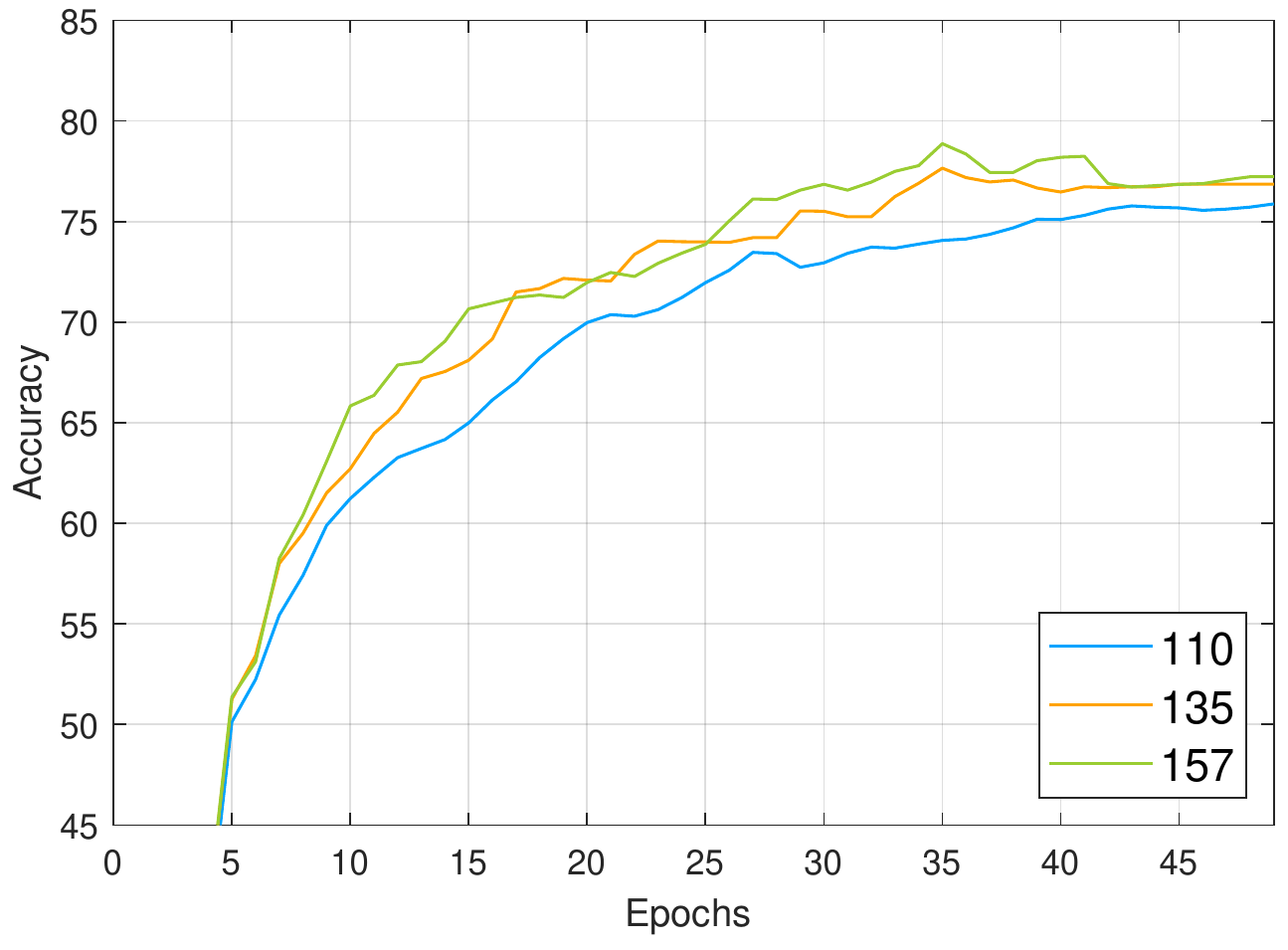}}
\end{minipage}
\caption{Average test accuracy of 20 experiments with random seeds during 50 epochs with different $\delta$ on Fashion MNIST and Rotated MNIST of DANN. $\delta = 157$ corresponds to \ours with only level one.}

\label{fig:exp2}
\end{figure}

For DANN, the training epochs are set to be 50. \cref{table:dann} shows the test accuracy of three sampling schemes on two datasets. Similar to MatchDG, while both $level_1$ and $level_2$ outperform $level_0$ on two datasets, $level_2$ gets relatively lower average accuracy than $level_1$ on Rotated MNIST and obviously outperforms $level_1$ on Fashion MNIST. A rational explanation is that Fashion MNIST in this experiment shows strong Osc which is weak on Rotated MNIST. Thus a small $\delta$ in level-two-sampling can efficiently alleviate the impacts from strong Osc on Fashion MNIST while it may lead to a certain degree of waste of training data on Rotated MNIST. \cref{fig:exp2} shows the average test accuracy for each epoch with different $\delta$ on these two datasets. On both datasets, while a smaller $\delta$ leads to slower growth in accuracy at the initial part of training, it helps get a more robust result, shrinking the gap between maximal accuracy and test accuracy. On Fashion MNIST, both $\delta=75$ and $\delta=115$ outperform training on all data batches, yet $\delta=157$ gets the best result on Rotated MNIST.

\begin{figure}[tbp]

\begin{minipage}[b]{.48\linewidth}
    \centering
    \subfloat[][Fashion MNIST]{\label{dann_fm_zhuzhuang}\includegraphics[width=6cm]{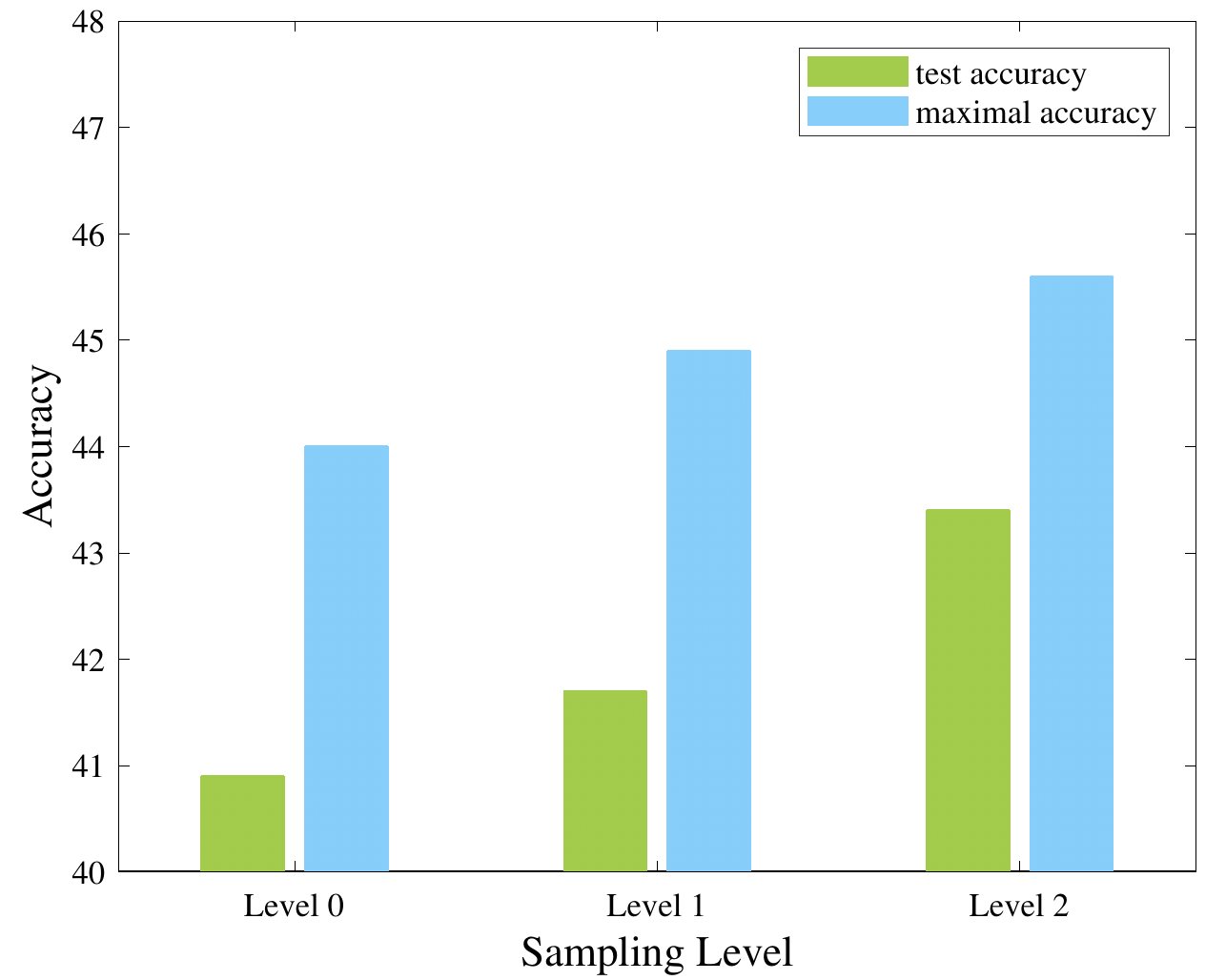}}

\end{minipage} 
\begin{minipage}[b]{.48\linewidth}
    \centering
    \subfloat[][Rotated MNIST]{\label{dann_rm_zhuzhuang}\includegraphics[width=6cm]{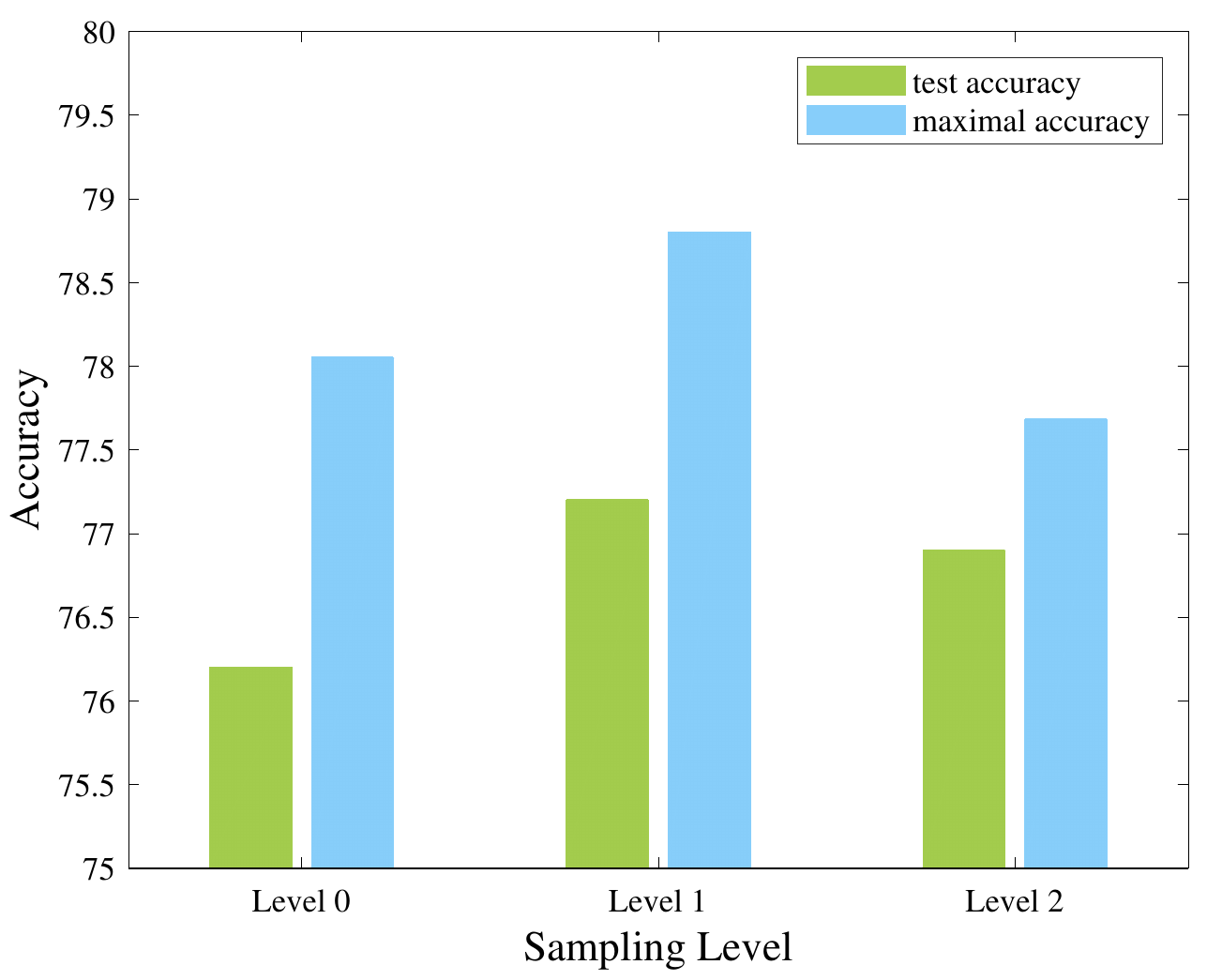}}
\end{minipage}
\caption{Average test accuracy and maximal accuracy of 30 times of experiments under different sampling levels on Fashion MNIST and Rotated MNIST of DANN.}
\label{fig:dann}
\end{figure}

\cref{fig:dann} shows the average maximal accuracy and test accuracy under different sampling levels of 30 experiments with random seeds. Similar to \cref{matchdg}, $level_1$ and $level_2$ enhance test accuracy and $level_2$ obviously shrinks the gap between test accuracy and maximal accuracy, showing that it helps get a more robust model.

\begin{table}[tbp]
\centering
\caption{ Average accuracy on Rotated MNIST and Rotated Fashion MNIST under three sampling schemes of DANN.}
\label{table:dann}
\begin{tabular}{cccc}
\toprule  
& $level_0$& $level_1$& $level_2$ \\
\midrule  
Rotated MNIST& 76.2& \textbf{77.2}& 76.9\\
Fashion MNIST& 40.9&41.7 & \textbf{43.4}\\

\bottomrule 
\end{tabular}
\end{table}

\subsection{FISH}
\begin{figure}[tbp]
\begin{minipage}[b]{.12\linewidth}
    \centering
   
    \includegraphics[width=1.3cm]{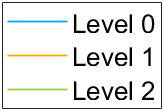}
\end{minipage}
\begin{minipage}[b]{.42\linewidth}
    \centering
    \subfloat[][Fashion MNIST]{\label{fish_fm}\includegraphics[width=5.5cm]{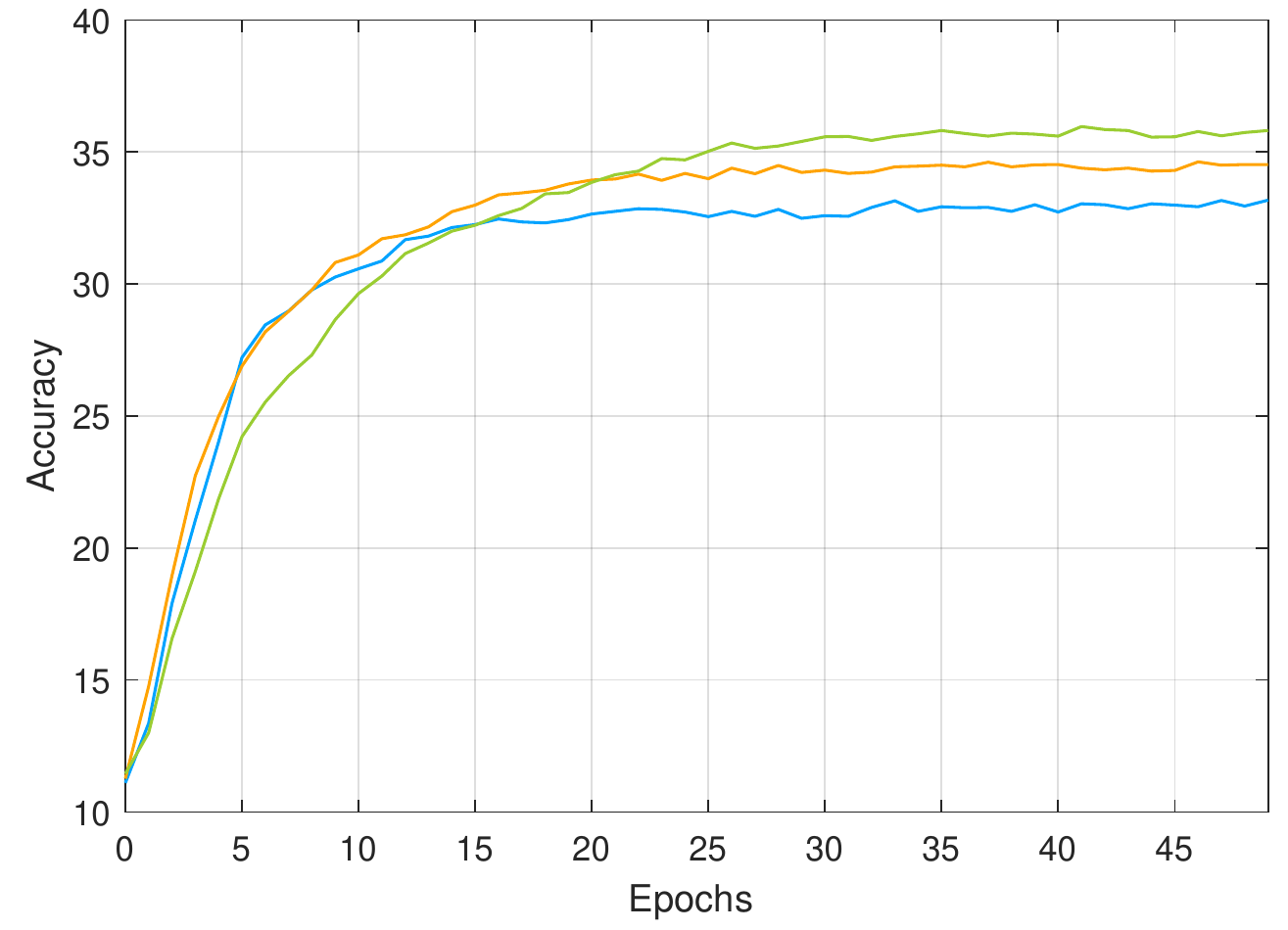}}

\end{minipage} 
\begin{minipage}[b]{.42\linewidth}
    \centering
    \subfloat[][Rotated MNIST]{\label{fish_rm}\includegraphics[width=5.5cm]{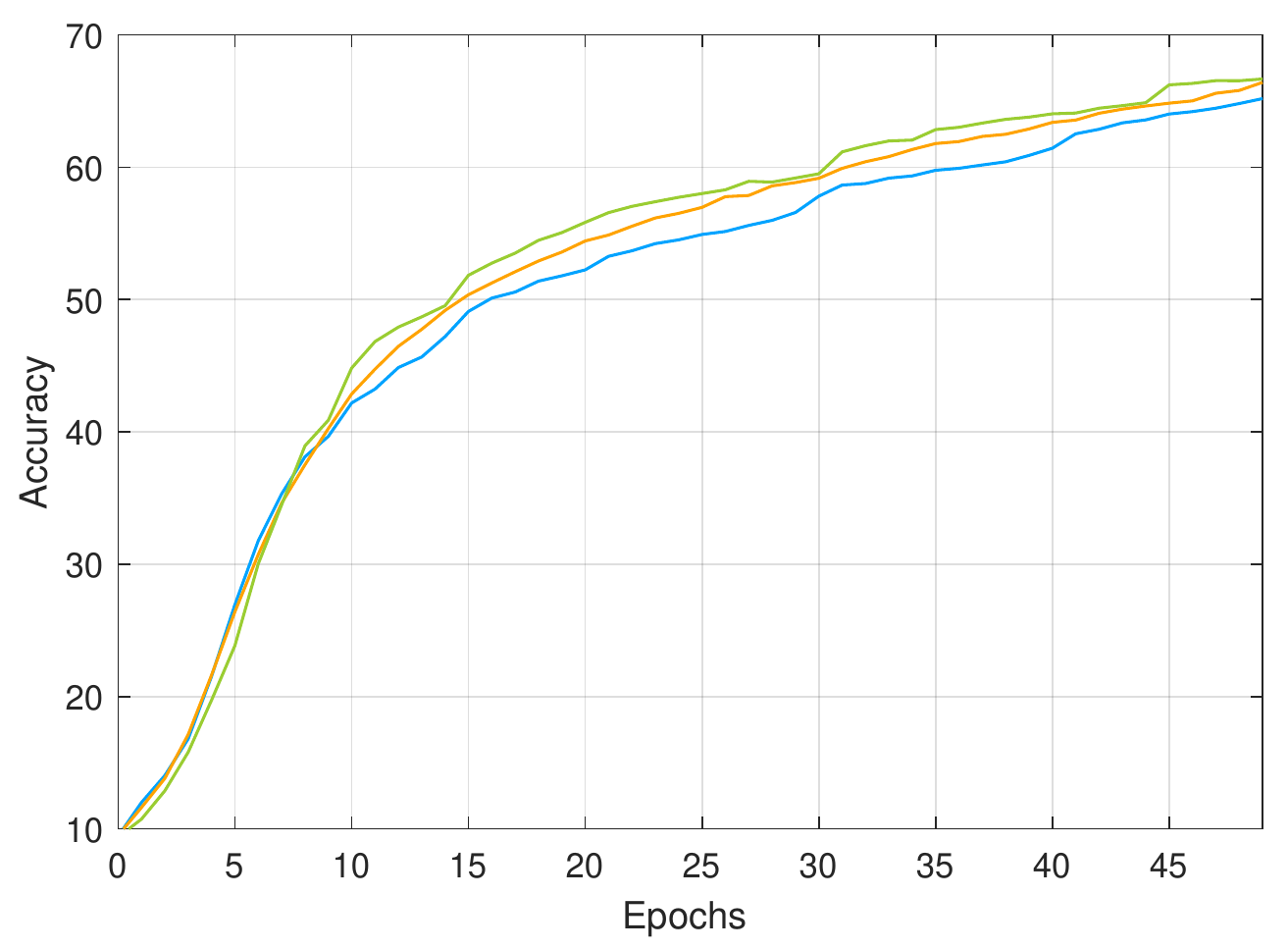}}
\end{minipage}

\caption{Average test accuracy of 10 experiments with random seeds during 50 epochs with different sampling schemes of FISH. Here we slightly abuse epoch to mean the time we obtain test accuracy.}
\label{fig:fish}
\end{figure}

For FISH, the training epochs are set to be 5. Each epoch contains 300 iterations and we observe test accuracy every 30 iterations. Unlike MatchDG and DANN, fish needs to sample domains in each iteration instead of training on one list of domains. Sampling domains in each iteration will result in great computational overhead compared to randomly sampling. Thus we just sample 30 domain lists containing diverse domains using level-one-sampling of \ours and repeatedly train the model on these domain lists(one list for one iteration) for $level_1$. As for $level_2$, we further utilize level-two-sampling to sample data batches of each domain in the domain lists for training. \cref{table:fish} shows the test accuracy. $level_2$ further enhances performance by $level_1$, and both of them apparently outperform $level_0$.\\
\cref{fig:fish} shows average test accuracy for each epoch with different sampling schemes on Fashion MNIST and Rotated MNIST. On both datasets, while $level_2$ leads to slower growth in accuracy at the initial part of training because of using a smaller number of batches, it keeps outperforming $level_1$ and $level_0$ at later epochs. $level_1$ also shows better performance than $level_0$.

\begin{table}[tbp]
\centering
\caption{ Average test accuracy of 10 experiments with random seeds on Rotated MNIST and Rotated Fashion MNIST under three sampling schemes of FISH. }
\label{table:fish}
\begin{tabular}{cccc}
\toprule  
& $level_0$& $level_1$& $level_2$ \\
\midrule  
Rotated MNIST& 65.2& 66.5& \textbf{66.6}\\
Fashion MNIST& 33.2&34.5 & \textbf{35.8}\\

\bottomrule 
\end{tabular}
\end{table}

% \subsubsection{iwildcam}
\subsection{Experiments on iwildcam}
The data of iwildcam is extremely unbalanced, while part of the domains contain less than 20 photos, some domains contain over 2000 ones.
In the original experiments of \cite{shi2021gradient}, iwlidcam is divided into batches in each domain. FISH samples a certain number of batches from different domains for training in each iteration. The sampling probability of one batch in a domain is proportional to the number of batches left in this domain. This sampling scheme is taken as $level_0$ here and the result of $level_0$ is taken from \citep{shi2021gradient}. In each iteration, $level_1$ samples the most diverse batches based on DPP using \invdann, $level_2$ samples some batches in the level-one-sampling and among them further selects a subset of batches in the level-two-sampling. Under the same setting in the original experiments, the results on iwildcam of FISH are shown in \cref{table:iwildcam} .\\
\begin{table}[tbp]
\centering
\caption{ Macro F1 score of FISH on iwildcam under three sampling schemes}
\label{table:iwildcam}
\begin{tabular}{cccc}
\toprule  
& $level_0$& $level_1$& $level_2$ \\
\midrule  
Iwildcam & 22.0&22.8 & \textbf{23.4}\\

\bottomrule 
\end{tabular}
\end{table}

Although \ours gets higher Macro F1 score, it leads to a much larger computational overhead since it needs to do sampling in each iteration. Moreover, for DANN and MatchDG, Macro F1 of diverse domains may be significantly lower than randomly sampled domains because of the unbalanced data, i.e., the diverse domains may contain much fewer data compared to the randomly sampled domains. It would be a significant future work to tackle the issues of extremely imbalanced data and computational overhead for algorithms that need to do sampling for multi-times.

\section{Conclusion}
Under the setting of large number of domains and domains with massive data points, we propose a diversity boosted two-level sampling algorithm named \ours to help sample the most informative subset of dataset. Empirical results show that \ours substantially enhances the out-of-domain accuracy and gets robust models against spurious correlations from both domain-side and object-side.

\clearpage   

\bibliography{reference}

\newpage
\appendix

\begin{center}
	\LARGE \bf {Appendix of \ours}
\end{center}

\etocdepthtag.toc{mtappendix}
\etocsettagdepth{mtchapter}{none}

\section{The Simulated Dataset}\label{table:feat}

\begin{table}[!htbp]
\centering
\caption{The simulated dataset of the toy example. From these 12 data points we sample 6 for training.}

\begin{tabular}{ccccccccccccc}
\toprule  
&$D_1$&$D_2$&$D_3$&$D_4$&$D_5$&$D_6$&$D_7$&$D_8$&$D_9$&$D_{10}$&$D_{11}$&$D_{12}$\\
\midrule  
$X_1$&0&0&0&0&0&0&0&0&0&1&1&1\\

$X_2$&0&0&0&0&0&0&1&1&1&0&1&1\\

$X_3$&0&0&0&0&1&1&1&1&1&1&1&1\\

$X_4$&0&0&0&1&0&1&1&1&0&0&1&1\\
\midrule 
$Y$ &0&0&0&0&0&0&1&1&1&1&1&1\\
\bottomrule 
\end{tabular}
\end{table}

\end{document}